\documentclass[runningheads]{llncs}

\usepackage{eccv}

\usepackage{eccvabbrv}

\usepackage{float}
\usepackage{graphicx}
\usepackage{booktabs}
\usepackage{adjustbox}
\usepackage{multicol}
\usepackage{multirow}
\usepackage{hyphenat}
\usepackage{xspace}
\usepackage{subcaption}

\usepackage[accsupp]{axessibility}  

\usepackage{hyperref}

\usepackage{orcidlink}

\newcommand{\datasetname}{PetFace\xspace}
\newcommand{\datasetnameshort}{\datasetname}

\makeatletter
\newcommand{\figcaption}[1]{\def\@captype{figure}\caption{#1}}
\newcommand{\tblcaption}[1]{\def\@captype{table}\caption{#1}}
\makeatother

\makeatletter
\newcommand{\printfnsymbol}[1]{%
        \textsuperscript{\@fnsymbol{#1}}%
}
\makeatother

\begin{document}

\title{\datasetname: A Large-Scale Dataset and Benchmark \\for Animal Identification} 

\titlerunning{\datasetname: A Large-Scale Dataset and Benchmark for Animal Identification}

\author{Risa Shinoda$^{1*}$ \orcidlink{0009-0006-3965-7933} \and
Kaede Shiohara$^{2*}$ \orcidlink{0009-0005-0603-3377}}

\authorrunning{R. Shinoda and K. Shiohara}

\institute{$^*$Equal contribution\\
$^1$Kyoto University, Japan \quad $^2$The University of Tokyo, Japan}
\maketitle

\begin{abstract}
Automated animal face identification plays a crucial role in the monitoring of behaviors, conducting of surveys, and finding of lost animals. Despite the advancements in human face identification, the lack of datasets and benchmarks in the animal domain has impeded progress. In this paper, we introduce the \datasetname dataset, a comprehensive resource for animal face identification encompassing 257,484 unique individuals across 13 animal families and 319 breed categories, including both experimental and pet animals. 
This large-scale collection of individuals facilitates the investigation of unseen animal face verification, an area that has not been sufficiently explored in existing datasets due to the limited number of individuals. 
Moreover, \datasetname also has fine-grained annotations such as sex, breed, color, and pattern.
We provide multiple benchmarks including re-identification for seen individuals and verification for unseen individuals. 
The models trained on our dataset outperform those trained on prior datasets, even for detailed breed variations and unseen animal families.
Our result also indicates that there is some room
to improve the performance of integrated identification on multiple animal families.
We hope the \datasetname dataset will facilitate animal face identification and encourage the development of non-invasive animal automatic identification methods. Our dataset and code are available at \url{https://dahlian00.github.io/PetFacePage/}.

\end{abstract}

\section{Introduction}
\label{sec:intro}
Animal identification plays a crucial role in animal studies and applications such as monitoring animal behavior, conducting habitat surveys, locating missing animals, and performing health checks. 
Traditional identification techniques, including ear tags, tattoos, ear punching, and toe clipping, continue to be utilized mainly for experimental animals and livestock~\cite{Dahlborn2013ReportOT}. 
However, given that these methods have the potential to cause stress and pain~\cite{welfare_tatto, mouse_welfare, LESLIE201086}, their use should be minimized to prioritize animal welfare. 
Therefore, there is a pressing need for the development and adoption of identification technologies that are not only effective and efficient but also minimally invasive, thereby mitigating the ethical concerns associated with traditional methods.
While advanced tools such as digital IDs~\cite{ROBERTS200618, Cloudtag} have been introduced, their applications involve a laborious process, \ie, attaching the devices to each animal individually. 
The process is costly and potentially stressful for the animals.
Moreover, these physical tags can identify only pre-defined individuals, which makes them impractical for use in real world scenarios.

In the human domain, digital face recognition is one of the effective approaches for the identification. 
It has been developed for use in smartphones, airport security, and systems for finding missing people. 
Therefore, the research community made great efforts to develop sophisticated deep learning-based face recognition models~\cite{arcface, center, cosface, sphereface, adaface, tripletnet}, empowered by large-scale datasets and benchmarks, \eg, ~\cite{lfw, frontal, agedb, ms1m}.

Despite the promise of human face recognition, the research progress towards automatic animal face individual recognition has been impeded, primarily because of the lack of extensive datasets and benchmarks for animal face recognition. 
Previous openly available datasets mostly include less than 100 individuals~\cite{macaques, horse,flicker_dog,ctaiczoo,gorilla_zoo}, which makes it far from generalized and discriminative identification models and precise evaluation for unseen individuals.

\begin{figure}[tb]
  \centering
  \includegraphics[height=4.8cm]{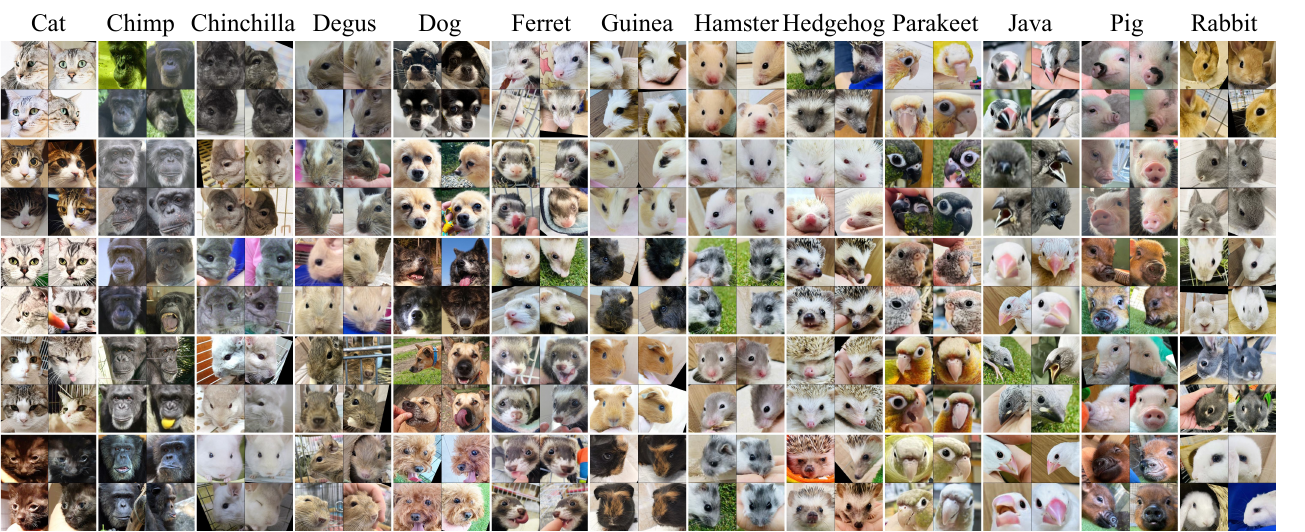}
  \caption{\textbf{Example images of our \datasetname.} 
  We introduce a large-scale animal face re-identification dataset \datasetname that include 257,484 unique individuals across 13 families and 319 breeds.
  From the left, the images represent Cat, Chimpanzee, 
  Chinchilla, Degus, Dog, Ferret, Guinea pig, Hamster, Hedgehog, Parakeet, Java sparrow, Pig, and Rabbit. 
  The four images enclosed within the white square's grid lines represent the same identity.}
  \label{fig:example}
  \vspace{-10pt}
\end{figure}

In this paper, we introduce a large-scale animal face recognition dataset called \datasetname that contains 257,484 individuals in total across 13 species with 319 breeds with 1,012,934 images. 
We show the example images of  our \datasetname in Fig~\ref{fig:example}.
The number of individuals in our dataset
is over 110 times that in the previous largest animal face dataset~\cite{DogFaceNet}. 
We sourced images and related information from the internet, with automated and manual filtering processes applied to ensure the dataset is not only large but also finely detailed and clean.
Moreover \datasetname has fine-grained annotations including sex, breeds, and colors and patterns of their skin, which allows further investigation for fine-grained recognition and evaluation.

\datasetname offers two benchmarks: one for recognizing known (seen) individuals and the other for recognizing unknown (unseen) ones. We also conduct the verification of the fine-grade breeds and unseen animal categories.

Our main contributions are as follows: 
(i) We establish a new dataset for animal face recognition called \datasetname, which contains a total of 257,484 individuals across 13 types of animal families and 319 breeds with fine-grade annotations including sex, breed, color of animals.
(ii) We set the benchmarks on recognizing known (seen) individuals and unknown (unseen) ones.
(iii) We show that the model trained using our dataset shows the generalization capabilities for unseen individuals and even for unseen animal categories.

\section{Related Work}

\begin{table}[htbp]
    \centering 
    \begin{tabular}{lcccc} 
    \toprule
    {Dataset} & {Family} & {\#Individuals} & {\#Images} & Fine-grained annotation\\
      \midrule
        CTai~\cite{ctaiczoo}&Chimp&78&5078&Age, Sex\\
        CZoo~\cite{ctaiczoo}&Chimp&24&2109&Age, Sex\\
        MacaqueFaces~\cite{macaques}&Monkey&34&6280&-\\
        BristolGorillas2020~\cite{gorilla_zoo}&Gorilla&7&5428&-\\
        DogFaceNet~\cite{DogFaceNet}&Dog&1393&8363&-\\
        Flickr-dog~\cite{flicker_dog}&Dog&42&374&-\\
        THODBRL2015~\cite{horse}&Horse&47&2820&Angle\\
        ZindiTurtleRecall~\cite{zindi}&Turtle&2265&12803&Angle\\
        SeaTurtleID2022~\cite{seaturtle2022}&Turtle&438&8729&Timestamp\\
        Ours &13 families&257,484&1,012,934&Sex, Breed, Color \& Pattern\\
      \bottomrule
      \end{tabular}
  \tblcaption{\textbf{Openly available animal face identification datasets.} Compared to other datasets, \datasetname has the largest number of individuals, and a wider range of families and breeds.}
  \label{tb:prior_research}
  \vspace{-10pt}
\end{table}
Building datasets and benchmarks is an important step in advancing animal re-identification through deep models. 
While earlier efforts have established the basics of animal face recognition, there is considerable potential for further development. 
Compared to the dataset for human face recognition~\cite{lfw, frontal, agedb, webface, ms1m}, those for animal faces~\cite{DogFaceNet,ctaiczoo,gorilla_zoo,macaques} have much fewer individuals.
The evaluation scenarios represented in these datasets often fall short of real-world applicability; most of the datasets focus on closed-set re-identification, rather than recognizing unseen individuals that is a critical requirement for practical applications. 
Our work enables the training and evaluation on a large number of individuals across a wider range of animal families and breeds.

\noindent \textbf{Human Face Identification} is the process of identifying an individual's identity using their unique facial characteristics.
The huge demands for individual identification have grown mainly in the human domain. 
In the identification of human faces, which is known as face recognition, considerable efforts have been expended in the research community~\cite{arcface, center, cosface, sphereface, adaface, tripletnet}. 
Because of their domain-agnostic frameworks, most of the state-of-the-art methods in face recognition can be exported into other domains, such as animal identification. 
Furthermore, these advancements have been supported by the introduction of large-scale datasets and benchmarks~\cite{lfw, frontal, agedb, webface, ms1m} that have played a crucial role in facilitating the exploration of powerful models. 
Motivated by this, we create the \datasetname dataset to fill the gap between human face recognition and animal face recognition.

\noindent \textbf{Animal Identification}
is the process of identifying an individual's identity using their unique body or facial characteristics, which is an important task across various scenarios, including monitoring animals, conducting habitat surveys, and finding missing animals.
With the advance of computer vision technology, various openly available datasets contribute to computer vision for animals~\cite{AerialCattle2017, ATRW,BirdIndividualID,Cows2021,Drosophila,FriesianCattle2017,Giraffes,MPDD,ndd20,noaa-right-whale-recognition,IPanda50,OpenCows2020,shinoda2024openanimaltracks,fishnet}.
Various methods are used to create the dataset, such as recording images~\cite{BirdIndividualID,GiraffeZebraID,MPDD,ndd20,noaa-right-whale-recognition} and videos~\cite{Zebrafish,ATRW,Drosophila,PolarBearVidID,IPanda50,Giraffes}, and using aerial image~\cite{AerialCattle2017,OpenCows2020,Cows2021}.
Recording individual information is labor intensive; therefore, most of the previous datasets contain only a limited number of individuals and are focused on one species.
Recently, WildlifeDatasets~\cite{WildlifeDatasets}, which gathers previously openly available datasets ~\cite{Zebrafish,AerialCattle2017,ATRW,BelugaID2022,BirdIndividualID,ctaiczoo,Cows2021,Drosophila,FriesianCattle2017,FriesianCattle2015,GiraffeZebraID,Giraffes,HappyWhale,HumpbackWhaleID,HyenaID2022,IPanda50,LeopardID2022,LionData,macaques,MPDD,ndd20,noaa-right-whale-recognition,OpenCows2020,PolarBearVidID,SealID,SeaStarReID2023,seaturtleid,SMALST,StripeSpotter,WhaleSharkID,zindi} combined, was introduced. 
This research creates the benchmarks using existing available datasets.

The advances in human face recognition raise an interest in animal face recognition.
One notable challenge is the variation in facial structures between animals and humans. 
This has led to studies like AnimalWeb~\cite{animalweb} and CatFLW~\cite{catflw} that propose specialized methods for animal facial key points detection.  
Automatic animal face identification has been studied, which can help humans to monitor animals~\cite{adaptive_pig,biometric_sheep,animalface_eccv20,Loos2013chimp,sheep_transformer}.
We review several publicly available face identification datasets in Table~\ref{tb:prior_research}.
Apes, such as Chimpanzees~\cite{ctaiczoo} and Gorillas~\cite{gorilla_zoo}, are one of the species that have been studied.  
However, as the dataset size is relatively limited, the datasets are insufficient to evaluate new unseen individual faces. 
In addition to primates, research on facial (head) identification has also extended to turtles, with the ZindiTurtleRecall~\cite{zindi} dataset offering a substantial number of images of individual turtles in a controlled environment. 
To address the limitations of in-the-wild data collection, the SeaTurtleID2022~\cite{seaturtle2022} dataset includes wild data, albeit with a reduced number of individuals, owing to the extensive effort required for data collection. 
Among datasets containing mammals, the DogFaceNet dataset~\cite{DogFaceNet} featuring 1,393 individual dogs stands out for its size.  
Nevertheless, the significant variation in appearance across dog breeds suggests that even this larger dataset may not be sufficiently comprehensive. 
In addition to the limited number of individuals, previous animal face datasets often focus on only single animal families. 
They can contribute to specific animal family research but can not explore animal identification across many families and breeds.

\section{\datasetname Dataset}
\datasetname is a large animal face identification dataset that  expands research on animal face recognition, which has been impeded by a scarcity of suitable datasets and benchmarks. 
This section details the construction of the \datasetname dataset, including our labor-efficient methods for collecting animal face images and a semi-automated filtering process to ensure quality fine-grained categorization and statistics of the dataset.
\subsection{Dataset Statistics}
\begin{figure}[H]
  \centering
  \includegraphics[height=14cm]{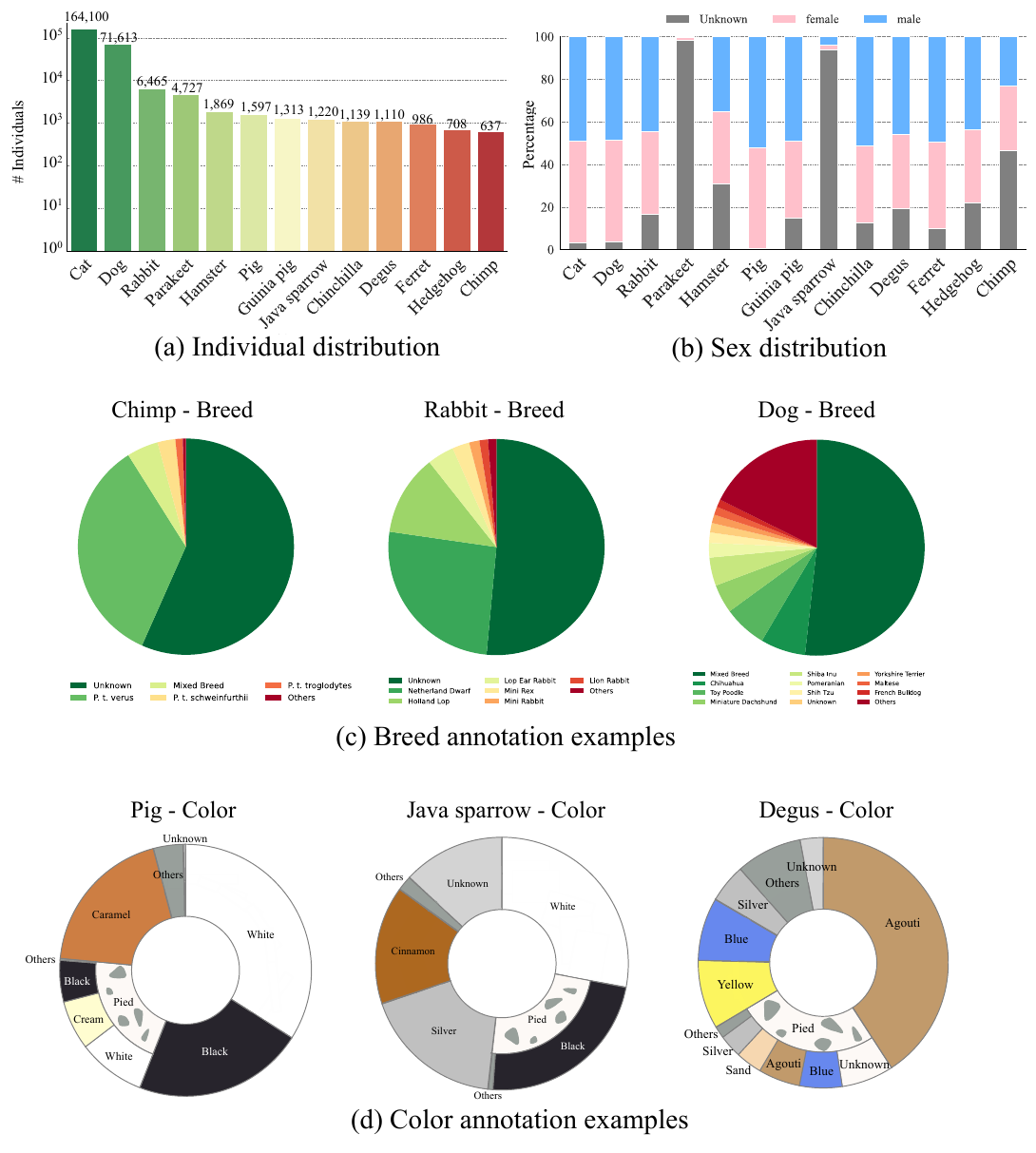}
  \caption{\textbf{Dataset distribution.} Each figure represents (a) the number of individuals in each animal family. (b) the sex distribution percentage by animal family. (c) examples of breed annotations. (d) examples of color annotations.}
  \label{fig:statistic}
\end{figure}
\label{sec:dataset}
The \datasetname dataset encompasses 1,012,934 images spanning 257,484 unique individuals. 
Detailed distributions of animal families are illustrated in Fig~\ref{fig:statistic}(a). 
Images are cropped to $224 \times 224$ pixels around their faces.
Our \datasetname also has fine-grained annotations.
Sex distribution across different animal families is depicted in Fig~\ref{fig:statistic}(b), with sex information available for 240,861 individuals, accounting for 94\% of the dataset.
The dataset includes annotations for 319 breeds, with examples of breed annotations shown in Fig~\ref{fig:statistic}(c). 
Furthermore, as detailed in Fig~\ref{fig:statistic}(d), the dataset provides in-depth color information through two-tier hierarchical annotations. 
Please see the supplemental for the detailed information per each animal family. 

\subsection{Data Sourcing}
Collecting images of animal faces via photography is labor and time intensive, which impedes the creation of large-scale datasets. 
In contrast, the human face recognition field benefits greatly from the availability of images sourced from the Internet. 
In this section, we outline our approach to assembling the collection of animal face images through the Internet, where each is associated with unique individual identifiers.

\noindent \textbf{Curation of images.}
In contrast to the human domain, acquiring multiple images for individual animals is more challenging. 
Unlike human datasets where a large number of celebrity images can be readily sourced, animal images require alternative approaches for curation. 
We utilized two primary sources: (i) pet shops' websites and (ii) animal adoption websites. 
The advantage of pet shops lies in their provision of high-quality, diverse images, capturing individuals from various angles and offering detailed information about each animal, including color, sex, and specific breed details. 
On the other hand, animal adoption sites offer images set against a variety of backgrounds and conditions that are often provided by pet owners, thus ensuring each individual is presented in a unique setting. 
These sources provide images that are highly suitable for animal recognition tasks and that are especially useful for recognizing animals in varied wild environments.
For Chimps, we additionally use images from the webpage of a collaborative research institution.

To ensure the dataset's quality, we were selective in choosing websites for curation. We only use the websites introducing each animal on one page to gain the individual IDs.
Aware of the potential for pet owners to upload the same images to multiple websites, we chose a single pet adoption website from each region (\eg, one per country) to minimize duplicates. 
For pet shops, we confirmed that the animals listed were unique to a certain shop to ensure the uniqueness of our dataset entries.
Using these sources, we collected 1,443,737 images from 325,420 individuals.

\subsection{Face Alignment and Filtering}
\begin{figure}[h]
  \centering
  \includegraphics[height=5.3cm]{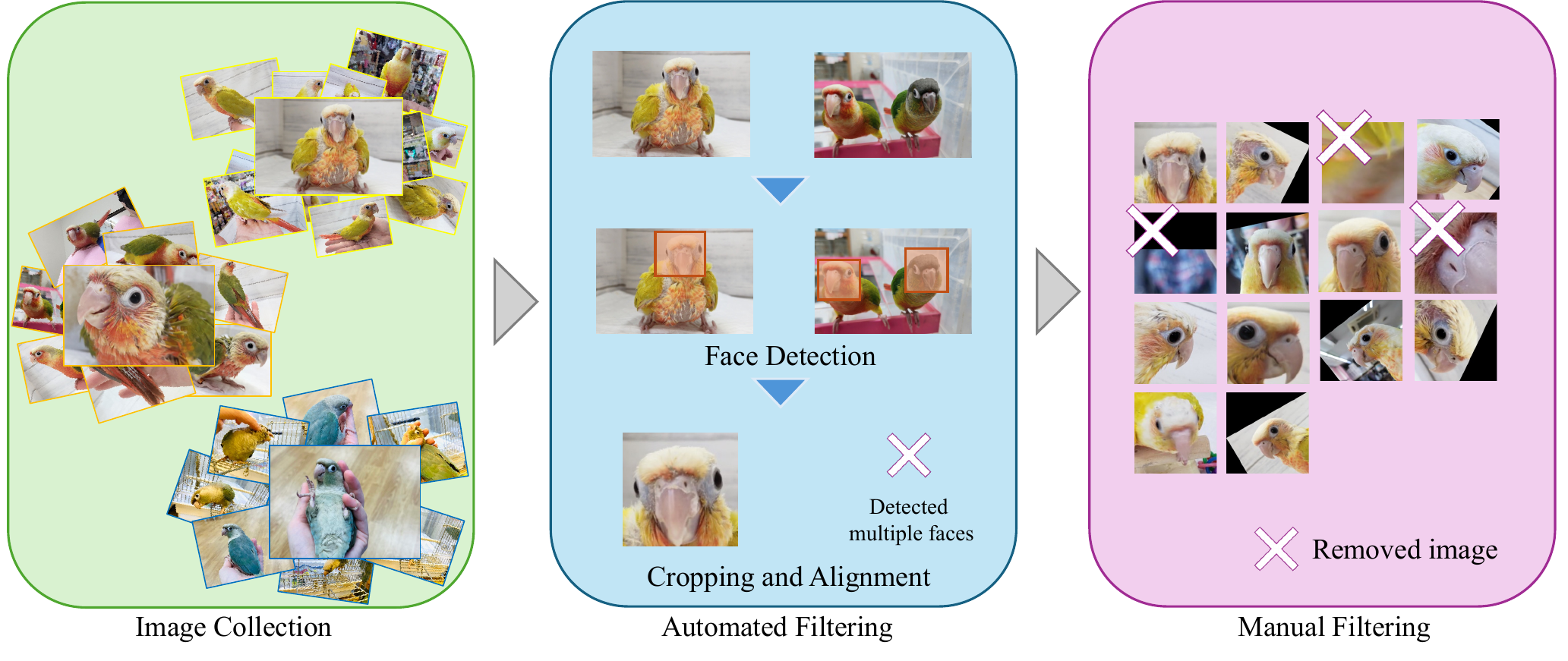}
  \caption{\textbf{Data filtering process.} We adopt a two-stage data filtering approach. Initially, images containing multiple faces are automatically removed. Subsequently, any images that do not depict animal faces or fail to meet alignment criteria are manually eliminated.}
  \label{fig:process}
\end{figure}
\noindent \textbf{Face Detection.}
We detect facial landmarks to align and crop the images. 
We adopt the AnyFace~\cite{anyface} that is trained on mixed face datasets including AnimalWeb~\cite{animalweb}.
Because the positions of animal facial parts are sometimes very different from those of humans, we use different reference points for different species in face alignment. 
After landmark detection, we select one frontal image as a reference for each species. 
We compute the average landmarks over all the landmarks aligned with the reference. 
Then, we define the target positions of the landmarks in the images; all the images are aligned to the target positions.

\noindent \textbf{Data Filtering.}
Fig~\ref{fig:process} shows the overview of our data filtering process.
Because the fully automatic face detection above sometimes fails, we adopt a two-stage data filtering process to filter out the following cases:
1) Images contain multiple animals simultaneously.
2) Images are unrelated to animals, such as advertisements. 
3) Images focus on non-animal elements, such as random patterns in backgrounds, toys, or people, rather than the animals themselves.
First, we automatically remove images where multiple faces are detected in an image.
Secondly, annotators manually assess all the images and remove those that do not depict the target animals or where the face alignment differs from our intended criteria.
Because checking whether the faces are properly aligned or not requires some expertise, all the images are filtered by the authors to ensure the quality of the dataset. 
This manual filtering process took about 100 man hours.
After this stage, we keep 1,012,934 images, which is approximately 70\% of the initial image number. 
Detailed distributions of animal families are illustrated in Fig~\ref{fig:statistic}(a).

\subsection{Fine-Grained Annotations}
Fine-grained categorization enhances its usefulness for downstream applications, \eg, the creation of challenging datasets based on animal's individual attributes.
However, collecting extensive individual animal data poses a major challenge in assembling large datasetse owing to the significant effort involved.
In response, we collect as much individual animal information corresponding to the images as possible.
Our annotations include sex, color, and specific breed details, depending on their availability on the source websites.
The annotation process involves: 1) extracting individual information displayed on websites alongside images, 2) refining raw text data to develop individual data tables, 3) manually verifying that all attributes are refined and free of unrelated information, and 4) manually annotating individuals with missing attributes on the website.

As a result, we include a sex category for all classes, with breed annotations for eight animal classes (Cat, Chimp, Dog, Guinea pig, Hamster, Parakeet, Pig, and Rabbit), and color annotations for eleven animal classes (Cat, Chinchilla, Degus, Ferret, Guinea pig, Hamster, Hedgehog, Parakeet, Java sparrow, Pig, and Rabbit). 
Note that the manual annotations are performed only on colors and patterns because images alone do not allow us to accurately determine an animal's sex or breed. Detailed distributions of sex per animal families are illustrated in Fig~\ref{fig:statistic}(b). In Fig~\ref{fig:statistic}(c) and (d), we illustrate the examples of breed annotations and color and pattern annotations.

\section{Experimental Methodology}
We introduce two principal evaluation protocols: 1) re-identification for seen faces, and 2) verification for unseen faces. To establish these benchmarks, we have curated two distinct types of test sets. Specifically, for unseen face verification, we carefully separate the dataset into training, validation, and test sets to prevent data leakage that can arise from shared backgrounds.
\subsection{Evaluation Protocols}
We adopt two types of evaluation. 

\begin{enumerate}
  \item Re-identification: This evaluation selects images from the test data that match the identities present in the training data. This procedure involves identifying test images that correspond to the same identities used during model training. The objective is to assess the model's ability to accurately recognize and associate test images with the correct identities from the training dataset. 
  \item Verification: This evaluation checks if the model can identify unseen faces. This means that we use different identities during the training and testing phases. 
  For each identity, we select one image that matches the identity and one image randomly chosen from a different identity. This process creates pairs of images where one pair consists of images from the same identity, and another pair consists of images from different identities. We then task the model with predicting whether the faces in each pair belong to the same identity or different identities. This approach allows us to comprehensively evaluate the model's ability to recognize and differentiate between individual identities. 
\end{enumerate}

\subsection{Data Split}
To establish benchmarks for seen individuals re-identification and unseen individuals verification, we use specific split protocols, creating two test sets designed for each task.
For seen individuals re-identification, the process involves verifying images of the same identities used in training; thus, we select test images from the same individuals for the training set. We, therefore, ensure that each individual is represented by at least three photos \ie, a minimum of two for training and one for testing.
For unseen individuals verification dataset, we ensure that it contains no images from sources that are also represented in the training or validation sets per animal family to prevent bias from similar environmental conditions. 
We divide the dataset into training, validation, and each testing sets following a 7:1:2 ratio as closely as possible, given the outlined criteria. 

\subsection{Models}
We train state-of-the-art models based on deep neural networks on our \datasetname to build the benchmarks. 
For the re-identification task, we approach the training and testing phases as classification problems; an individual is assigned to a class. 
On the other hand, for the verification task, we train models in the same manner as for the re-identification task but evaluate the models by computing the cosine similarity between pairs of images to be identified if they are the same individual.
Specifically, we focus more on the loss functions that are crucial for identification tasks than network architectures.
We refer to three important loss functions in addition to the basic Softmax-based classification model:

\noindent \textbf{Triplet loss}~\cite{tripletnet} is designed to take a triplet of samples, \ie, an anchor $x_{a}$, a positive $x_{p}$~(another image of the same identity as the anchor), and a negative $x_{n}$~(an image of a different identity) - and learn embeddings in such a way that the $x_{p}$ is closer to the $x_{a}$ than the $x_{n}$ by a margin. This loss function is particularly beneficial for face Re-ID as it directly targets the relative distances between different and same identity pairs, encouraging the model to learn a feature space where embeddings of the same identity are clustered together while being far from other identity clusters. 

\noindent \textbf{Center loss}~\cite{center} works alongside Softmax Loss to enhance the discriminative power of the learned features. While Softmax Loss focuses on inter-class separability, Center Loss aims to minimize the intra-class variations. It does this by penalizing the distance between the deep features of each class and their corresponding class center. Center Loss ensures that the embeddings of the faces of the same individual are closer together, thus making the feature distribution more compact for each identity. 

\noindent \textbf{ArcFace loss}~\cite{arcface} introduces an angular margin between classes to enforce a discriminative feature space. 
It modifies the Softmax loss by adding a margin penalty to the angle between the feature vector and the corresponding class center in the angular space. This angular margin encourages models to learn more distinguishable embeddings to separate classes.

We use ResNet-50~\cite{resnet}, which we found performs better than recent Transformer\hyp{}based models in Table~\ref{tb:backbone}, as our base backbone for all applied loss functions to simplify our experiments and make them easier to grasp. 

In addition to the comparison on the models trained on \datasetnameshort, we evaluate state-of-the-art models trained on other datasets including:

\noindent \textbf{ImageNet}~\cite{imagenet} is a conventional image classification dataset including 1000 general object classes. 
We use the ResNet-50 trained on the dataset.

\noindent \textbf{CLIP}~\cite{clip} learns semantic relationships between images and texts in a cross-modal contrastive learning manner empowered by web\hyp{}scale 
image\hyp{}caption datasets.
We use ResNet-50 backbone.

\noindent \textbf{MegaDescriptor}~\cite{WildlifeDatasets} is a state-of-the-art model for animal identification trained on a unified animal identification dataset
including 33 existing animal re\hyp{}identification datasets~\cite{Zebrafish,AerialCattle2017,ATRW,BelugaID2022,BirdIndividualID,ctaiczoo,Cows2021,Drosophila,FriesianCattle2017,FriesianCattle2015,GiraffeZebraID,Giraffes,HappyWhale,HumpbackWhaleID,HyenaID2022,IPanda50,LeopardID2022,LionData,macaques,MPDD,ndd20,noaa-right-whale-recognition,OpenCows2020,PolarBearVidID,SealID,SeaStarReID2023,seaturtleid,SMALST,StripeSpotter,WhaleSharkID,zindi}.
We use the officially provided SwinTransformer-B~\cite{swin} backbone.

\section{Experimental Result}
\subsection{Benchmark on Animal Face Re-identification}
\label{sec:reid}
We show the re-identification results in terms AUC in Table~\ref{tb:reidentification}. 
We train models with the baseline loss functions independently on each class. 
We can see that ArcFace performs consistent results, 51.23\% of average accuracy, compared to the other loss functions (41.88\% for Softmax and 9.81\% for Center).
On the other hand, Center loss does not learn sufficient discriminative features for animal face re-identification, especially on Cat and Dog where the number of the test classes are 113,592 and 46,755, respectively.
These results encourage the community to explore more effective representation learning methods for re-identification tasks. 

Moreover, motivated by MegaDescriptor, we train an ArcFace model on the entire \datasetname, denoted as \textit{Joint-Trained on PetFace} in the table.
We observe that the joint-training strategy provides some improvements, \eg, from 54.29\% to 70.30\% for Cat and from 29.08\% to 41.49\% for Pig although in some classes the results get worse, \eg, from 43.27\% to 34.30\% for Chimp and from 62.19\% to 44.78\% to Parakeet. 
This result indicates that there is some room to improve the performance of integrated identification on imbalance and wide-range datasets.

\begin{figure*}[t]
  \centering
  \begin{minipage}{.32\textwidth}
    \centering
    \includegraphics[height=3.5cm]{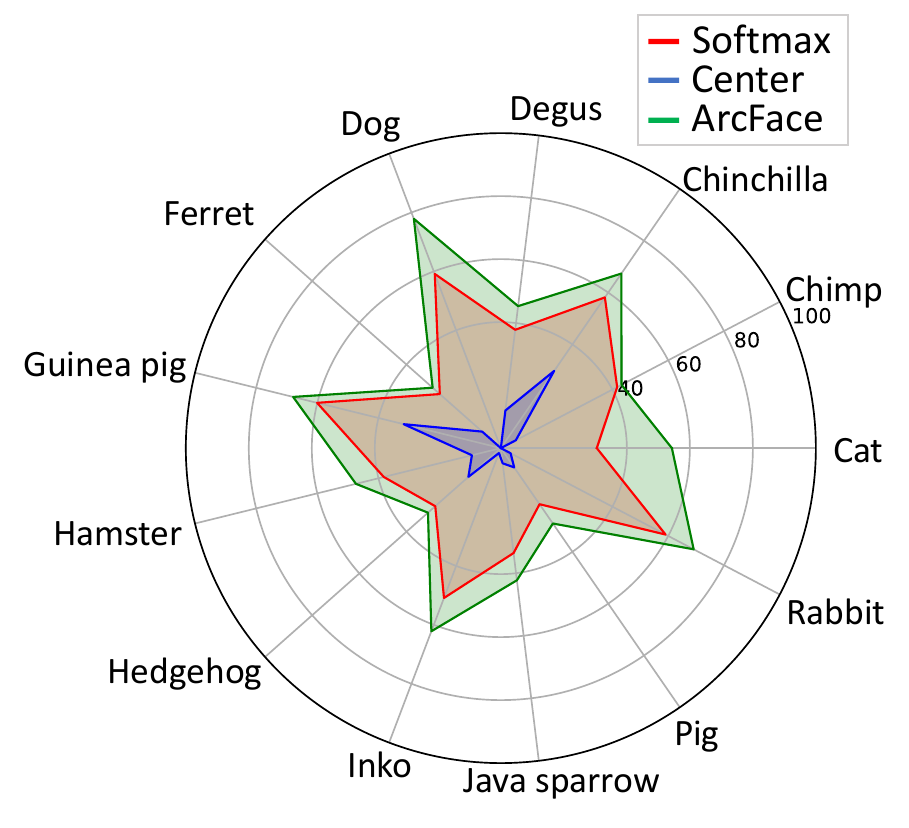}
    \subcaption{Top-1}
  \end{minipage}
  \begin{minipage}{.32\textwidth}
    \centering
    \includegraphics[height=3.5cm]{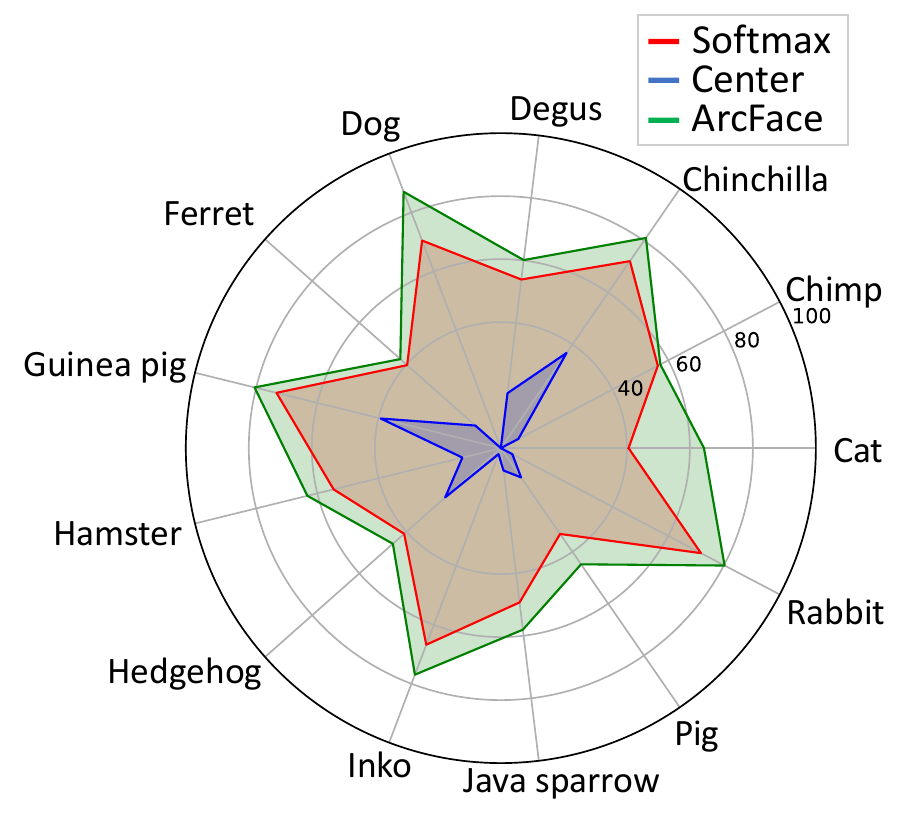} 
    \subcaption{Top-3}
  \end{minipage}
  \begin{minipage}{.32\textwidth}
    \centering
    \includegraphics[height=3.5cm]{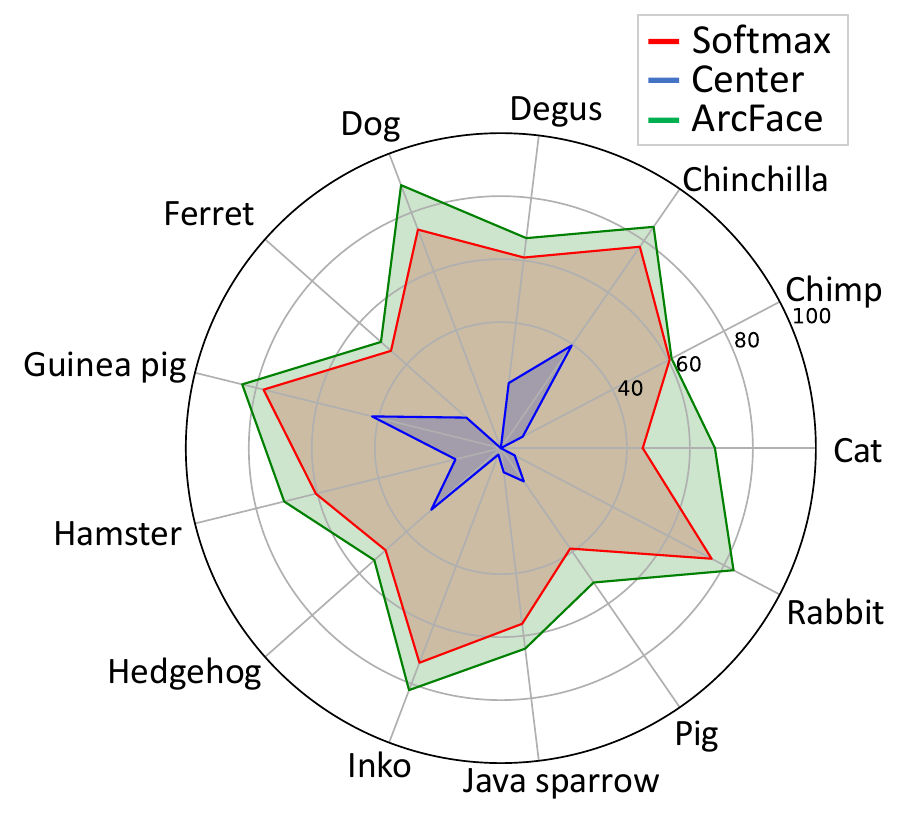} 
    \subcaption{Top-5}
  \end{minipage}
  \caption{\textbf{Top-k accuracy~($k=1,3,5$).}
  ArcFace consistently shows the best performance consistently.}
  \label{fig:topk}
\end{figure*}

In Fig~\ref{fig:topk}, we display the Top-$k$~($k=1,3,5$) accuracy metrics. While accuracy naturally increases with larger $k$ values, the relative ranking of accuracy across different animal families remains largely consistent.
ArcFace maintains the best performance across these evaluations.
\begin{table*}[t]
    \centering
    \begin{adjustbox}{width=1.0\linewidth}
    \begin{tabular}{lccccccccccccc|c} \toprule
      \multirow{2}{*}{Method} & \multicolumn{13}{c}{Top-1 Accuracy (\%)}& \multirow{2}{*}{Avg}\\
      \cmidrule(){2-14} 
      &Cat& Chimp &Chinchilla & Degus & Dog& Ferret&Guinea &Hamster &Hedgehog &Parakeet &Java sparrow& Pig & Rabbit\\
      \midrule
        \multicolumn{4}{l}{\textit{Trained on \datasetnameshort}}\\
        Softmax & 30.46 & 41.70 & 58.13 & 37.84 & 59.14 & 25.88 & 60.07 & 38.27 & 27.81 & 50.88 & 33.55 & 21.64 & 59.05 & 41.88 \\
        Center & 0.00 & 5.38 & 29.76 & 12.03 & 0.00 & 7.91 & 31.77 & 9.46 & 13.76 & 1.68 & 4.86 & 7.44 & 3.51 & 9.81 \\
        Arcface & 54.29 & \textbf{43.27} & 67.34 & 45.41 & \textbf{77.86} & 28.92 & 67.90 & 47.37 & 30.90 & \textbf{62.19} & \textbf{42.27} & 29.08 & \textbf{69.13} & 51.23 \\
        \midrule
        \multicolumn{4}{l}{\textit{Joint-Trained on \datasetnameshort}}\\
        ArcFace & \textbf{70.30} & 34.30 & \textbf{69.86} & \textbf{56.08} & 68.75 & \textbf{46.12} & \textbf{68.66} & \textbf{54.33} & \textbf{44.38} & 44.78 & 34.55 & \textbf{41.49} & 65.75 & \textbf{53.80} \\  
      \bottomrule
    \end{tabular}
    \end{adjustbox}
  \caption{\textbf{Result on animal re-identification.}
  ArcFace outperforms the other loss functions.
  The jointly trained model on our \datasetname shows the best average top-1 accuracy.}
  \label{tb:reidentification}
\end{table*}

\begin{table*}[t]
    \centering
    \begin{adjustbox}{width=1.0\linewidth}
    \begin{tabular}{lccccccccccccc|c} \toprule
      \multirow{2}{*}{Method} & \multicolumn{13}{c}{AUC (\%)} & \multirow{2}{*}{Avg}\\
      \cmidrule(){2-14} 
      &Cat& Chimp &Chinchilla & Degus & Dog& Ferret&Guinea pig&Hamster &Hedgehog &Parakeet &Java sparrow& Pig & Rabbit&\\
      \midrule
        \multicolumn{4}{l}{\textit{Trained on \datasetnameshort}}\\

        Softmax & 97.97 & \textbf{85.22} & 84.44 & 86.19 & 98.98 & 83.15 & 95.30 & 87.73 & 87.27 & 98.39 & 84.67 & 87.09 & 98.53 & 90.38 \\
        Center & 50.20 & 81.60 & 80.96 & 81.34 & 59.64 & 78.77 & 92.10 & 83.59 & 85.26 & 96.91 & 80.54 & 82.58 & 97.01 & 80.81 \\
        Triplet & 96.94 & 77.10 & 76.12 & 72.80 & 97.97 & 79.48 & 83.37 & 80.08 & 81.47 & 91.56 & 77.86 & 75.72 & 94.74 & 83.48 \\
        Arcface & 97.71 & 83.76 & 87.70 & \textbf{87.61} & \textbf{99.45} & 86.24 & \textbf{96.03} & 90.32 & 86.13 & \textbf{98.63} & 84.45 & 89.88 & \textbf{99.01} & 91.30 \\
        \midrule
        \multicolumn{4}{l}{\textit{Joint-Trained on \datasetnameshort}}\\
        ArcFace & \textbf{98.04} & 83.20 & \textbf{89.63} & 86.79 & 99.01 & \textbf{88.44} & 94.14 & \textbf{92.50} & \textbf{87.86} & 98.40 & \textbf{89.38} & \textbf{92.38} & 98.45 & \textbf{92.17} \\
        \midrule
        \multicolumn{4}{l}{\textit{Trained on Other Dataset}}\\
        ImageNet & 81.58 & 73.67 & 70.80 & 70.69 & 97.18 & 67.95 & 85.19 & 74.13 & 75.01 & 90.45 & 79.49 & 82.42 & 86.23 & 79.60 \\
        CLIP & 85.10 & 70.70 & 73.42 & 74.66 & 91.86 & 70.89 & 80.67 & 73.91 & 77.65 & 83.19 & 76.52 & 80.02 & 83.55 & 78.63\\
        MegaDescriptor & 87.30 & 83.01 & 77.85 & 77.79 & 93.75 & 76.53 & 88.36 & 77.91 & 78.74 & 89.76 & 78.64 & 88.08 & 90.38 & 83.70  \\
      \bottomrule
    \end{tabular}
    \end{adjustbox}
  \caption{\textbf{Result on animal verification.}
  Jointly trained on our \datasetname shows the best performance. We also compared to the trained models on other dataset.}
  \label{tb:verification}
\end{table*}

\begin{figure*}[t]
  \centering
  \begin{minipage}{.32\textwidth}
    \centering
    \includegraphics[height=3cm]{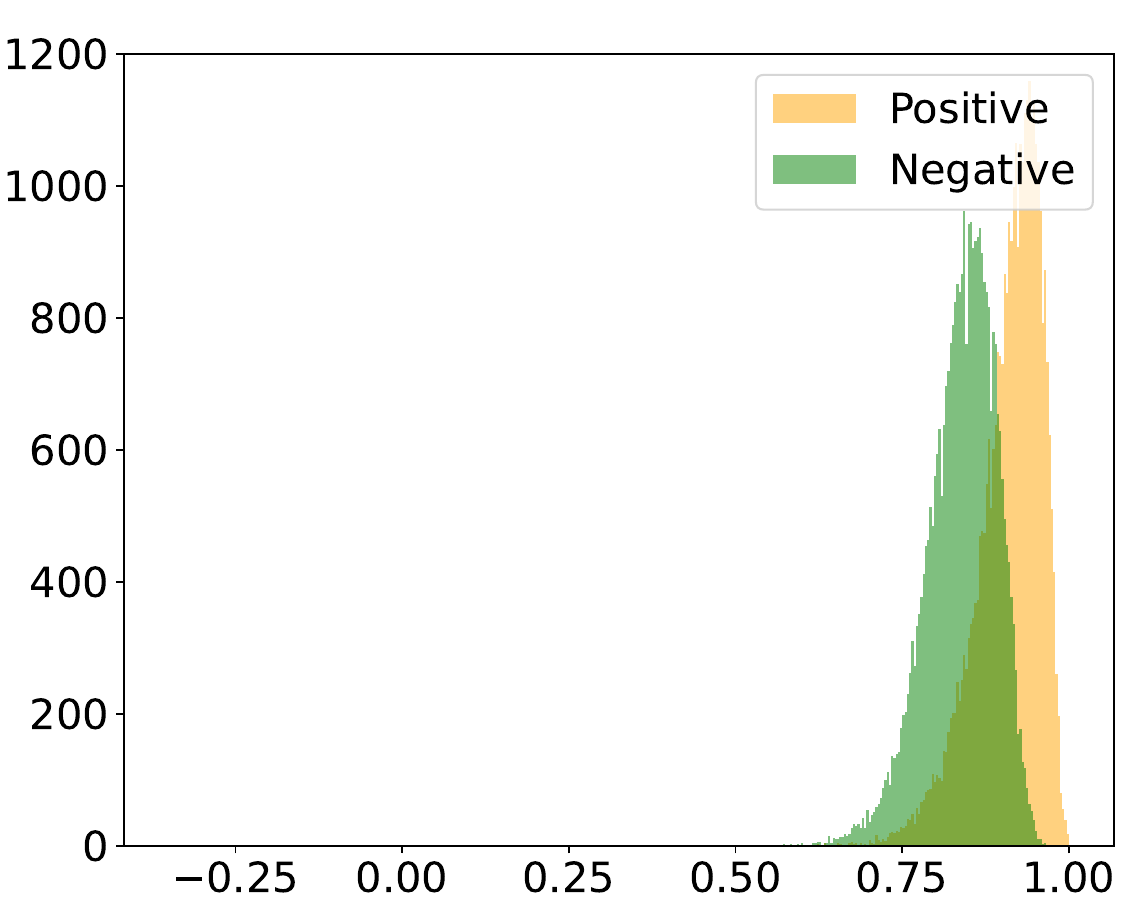}
    \subcaption{CLIP}
  \end{minipage}
  \begin{minipage}{.32\textwidth}
    \centering
    \includegraphics[height=3cm]{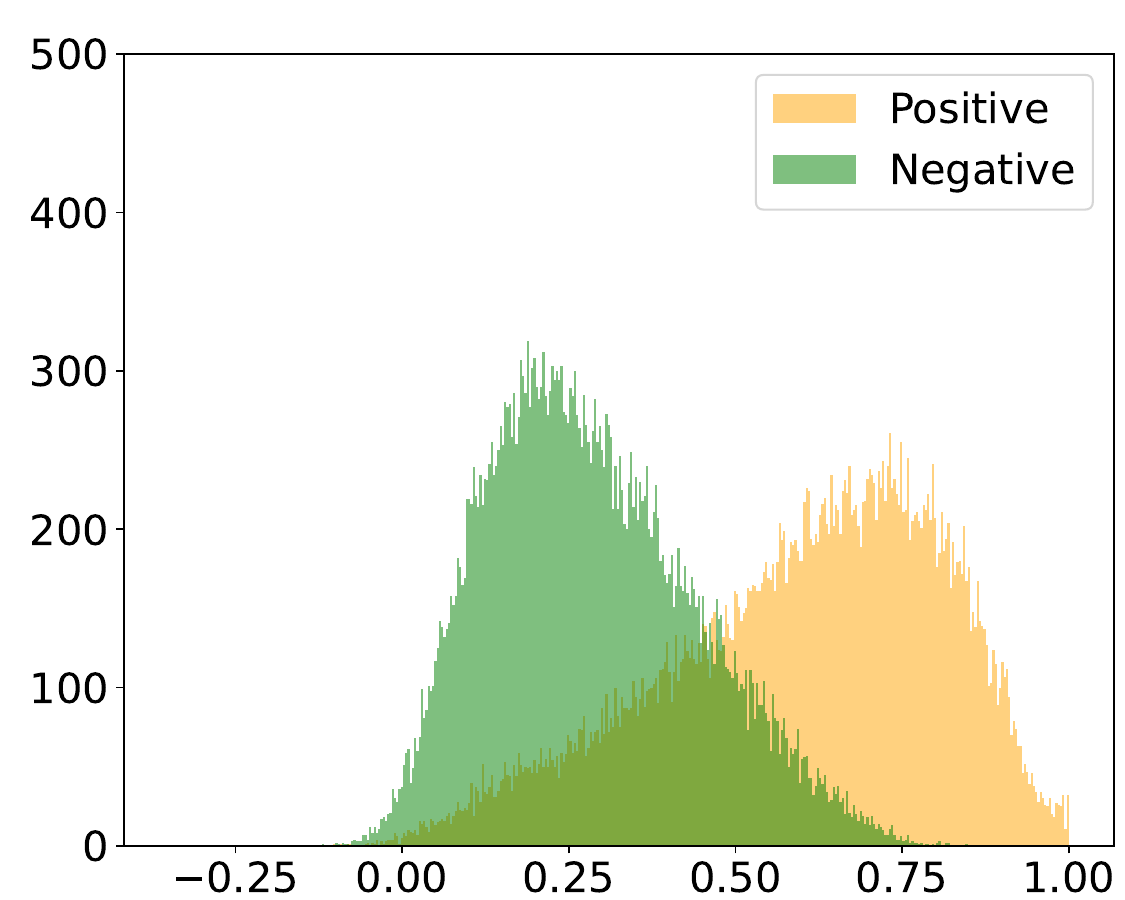} 
    \subcaption{MegaDescriptor}
  \end{minipage}
  \begin{minipage}{.32\textwidth}
    \centering
    \includegraphics[height=3cm]{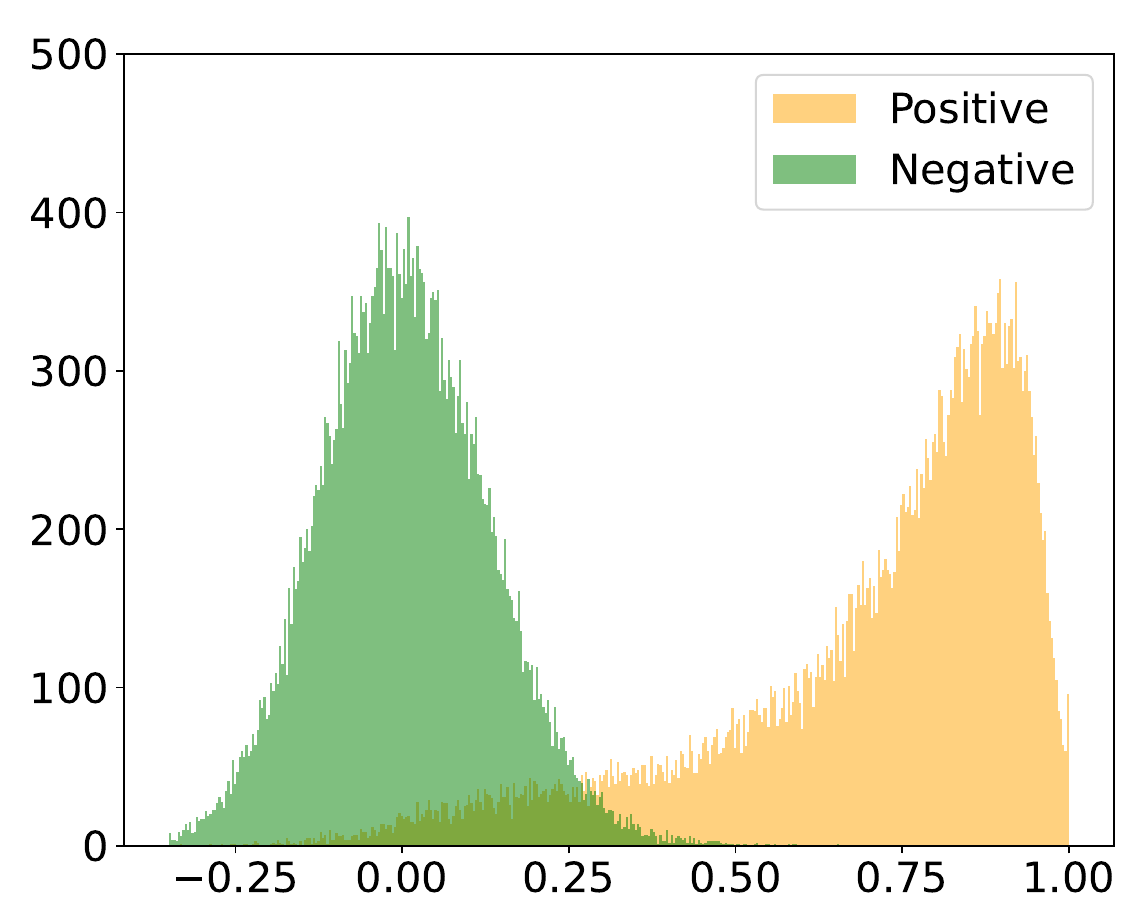} 
    \subcaption{Ours}
  \end{minipage}
  \caption{\textbf{Similarity distributions of (a) CLIP, (b) MegaDescriptor, and (c) our joint-trained model on cats.} The horizontal axis represents similarity and the vertical axis represents frequency. The model trained on \datasetname shows the most distinct separation between positive and negative samples.
  }
  \label{fig:similarity}
\end{figure*}

\subsection{Benchmark on Animal Face Verification}
Next, we show the verification results in Table~\ref{tb:verification}. 
We use the trained models in Table~\ref{tb:reidentification} and evaluate them on unseen individuals.
Similar to our findings in the re-identification task, ArcFace proved to be effective, achieving the best results with an average AUC of 92.17\% when jointly trained on the \datasetname dataset. 
Similar to face re-identification in Sec~\ref{sec:reid}, in some classes, the results get worse when jointly trained across animal families \eg, Chimp, Degus, Dog, Guinea pig, Parakeet, and Rabbit.
We also examined models pre-trained on other datasets including ImageNet, CLIP and MegaDescriptor in order to make more comparisons. 
Among these models, MegaDescriptor, which trained on the animal re-identification dataset, showed the highest AUC of 83.70\%.
We show the similarity distribution on the our cat dataset in Fig~\ref{fig:similarity}.
Comparing the performance of CLIP, MegaDescriptor, and our joint-trained models, our findings indicate that our model achieves the most distinct separation between positive and negative samples. In contrast, the MegaDescriptor model displays a closer distribution of these samples, and with CLIP, the overlap between positive and negative distributions is significantly more pronounced.

\subsection{Comparison with Previous Datasets}
 \begin{table*}[h]
    \begin{adjustbox}{width=1.0\linewidth}
       \begin{subtable}{0.6\textwidth}
           \begin{tabular}{lcccc} \toprule
     \multicolumn{3}{c}{Train Set} & \multicolumn{2}{c}{Test Set AUC (\%)}\\
       \cmidrule(l{2pt}r{2pt}){1-3} \cmidrule(l{2pt}r{2pt}){4-5}
      Database&\#Ind.&\#Images&CTai&CZoo\\
      \midrule
        CTai&49&2857&  {\color{lightgray}75.80} & 69.31 \\ 
        CZoo&17&1323&66.76 & {\color{lightgray}80.49} \\ 
        Ours-Chimp&446&2679&  \textbf{67.33} & \textbf{71.27} \\ 
      \bottomrule
    \end{tabular}
           \caption{Chimp datasets.}
       \end{subtable}%
       \begin{subtable}{0.65\textwidth}
           \begin{tabular}{lcccc} \toprule
     \multicolumn{3}{c}{Train Set} & \multicolumn{2}{c}{Test Set AUC (\%)}\\
       \cmidrule(l{2pt}r{2pt}){1-3} \cmidrule(l{2pt}r{2pt}){4-5}
      Database&\#Ind.&\#Images&DogFaceNet&Flickr-dog\\
      \midrule
        DogFaceNet & 975 & 5879& {\color{lightgray}98.72} & 85.51 \\ 
        Flickr-dog & 30 &  265&94.98 & {\color{lightgray}87.77} \\ 
        Ours-Dog &46755&168348& \textbf{99.65} & \textbf{95.54} \\ 
      \bottomrule
    \end{tabular}
           \caption{Dog datasets.}
       \end{subtable}
       \end{adjustbox}
       \caption{\textbf{Comparisons with previous datasets for Chimp and Dog.} The results when the test sets are the same as the training datasets are excluded for fair comparisons.
       The models trained on our dataset outperform those trained on previous datasets.}
       \label{tb:comparison_with_previous_datasets}
   \end{table*}
We conduct cross-dataset evaluations where models are tested on different datasets other than the training datasets to demonstrate the generality of our dataset. 
Here, we compare our dataset with 1) CTai~\cite{ctaiczoo} and CZoo~\cite{ctaiczoo} for Chimpanzee and 2) DogFaceNet~\cite{DogFaceNet} and Flickr-dog~\cite{flicker_dog} for Dog. 
All images of the compared datasets are aligned in the same manner as ours for fair comparisons.
We split the identities of the compared datasets into training and test sets in a ratio of 7:3 and then uniformly pick the positive and negative pairs.
Because CTai, CZoo, and Flickr-dog have a few test identities, we each pick five positive and negative pairs for each test identity for robust evaluations.  

We show the verification results in Tables~\ref{tb:comparison_with_previous_datasets}(a) and (b).
We exclude the results trained on the same dataset as the test set denoted in {\color{lightgray}gray} for fair comparisons.
For the Chimpanzee datasets, the model trained on our dataset outperforms previous datasets. For case tested on CTai, our dataset outperforms CZoo (67.33\% vs. 66.76\%). For the other case tested on CZoo, our dataset also surpasses CTai (71.27\% vs. 69.31\%).
This result demonstrates that our Chimpanzee dataset is more effective and general than the previous datasets.
For the Dog datasets, owing to the significant scale of our dataset, the model trained on our dog dataset outperforms not only cross-dataset results but also in-dataset ones denoted in {\color{lightgray}gray}. 
These results support the quality of our \datasetnameshort dataset.

\subsection{Analysis}
\begin{table*}[t]
    \centering
    \begin{adjustbox}{width=1.0\linewidth}
    \begin{tabular}{lccccccccccccc|c} \toprule
      \multirow{2}{*}{Method} & \multicolumn{13}{c}{AUC (\%)} & \multirow{2}{*}{Avg}\\
      \cmidrule(){2-14} 
      &Cat& Chimp &Chinchilla & Degus & Dog& Ferret&Guinea pig&Hamster &Hedgehog &Parakeet &Java sparrow& Pig & Rabbit&\\
      \midrule
        ResNet-50 & \textbf{97.71} & \textbf{83.76} & \textbf{87.70} & \textbf{87.61} & \textbf{99.45} & \textbf{86.24} & \textbf{96.03} & \textbf{90.32} & 86.13 & \textbf{98.63} & 84.45 & \textbf{89.88} & \textbf{99.01} & \textbf{91.30} \\
        ViT-32-B & 94.44 & 78.60 & 71.13 & 82.69 & 93.39 & 68.17 & 83.21 & 77.89 & \textbf{87.77} & 95.07 & 80.62 & 75.31 & 95.18 & 83.35 \\
        Swin-B & 87.35 & 74.23 & 76.21 & 84.04 & 85.75 & 71.60 & 91.56 & 88.33 & 86.22 & 94.12 & \textbf{84.66} & 66.56 & 90.56 & 83.17 \\
        \bottomrule
    \end{tabular}
    \end{adjustbox}
  \caption{\textbf{Verification result on different backbones.}
  ResNet-50 achieves the best average AUC.}
  \label{tb:backbone}
\end{table*}
\noindent\textbf{Different networks.}
To examine the impact of different network architectures on the performance on \datasetnameshort, we additionally train two backbones, VisionTransformer-32-B~\cite{vit} and SwinTransformer-B~\cite{swin}, with the same training configuration as our base model (ResNet-50) and report the verification results in Table~\ref{tb:backbone}.
We observe that ResNet-50 outperforms the additional transformer-based models in most cases and achieves the best average AUC of 91.30\%.

\begin{figure}[t]
    \begin{tabular}{cc}
        \begin{minipage}{.5\textwidth}
            \centering
            \includegraphics[width=1.0\linewidth]{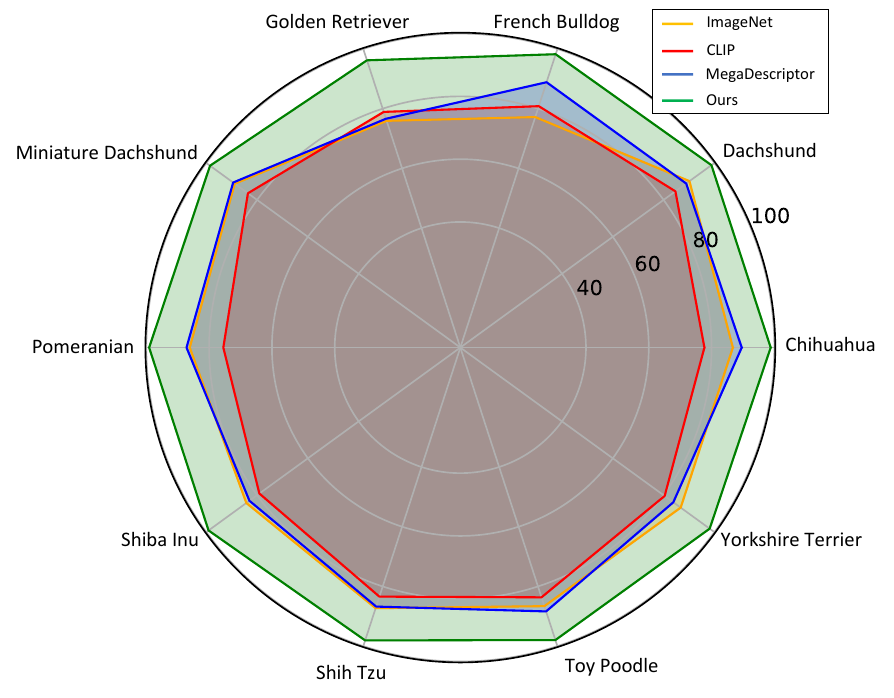}
            \caption{\textbf{Fine-grained dog verification.}
            Trained on our \datasetname Dog achieves the best performance consistently.}
            \label{fig:fine_dog}
        \end{minipage}
        \begin{minipage}{.47\textwidth}
            \centering
            \includegraphics[width=1.0\linewidth]{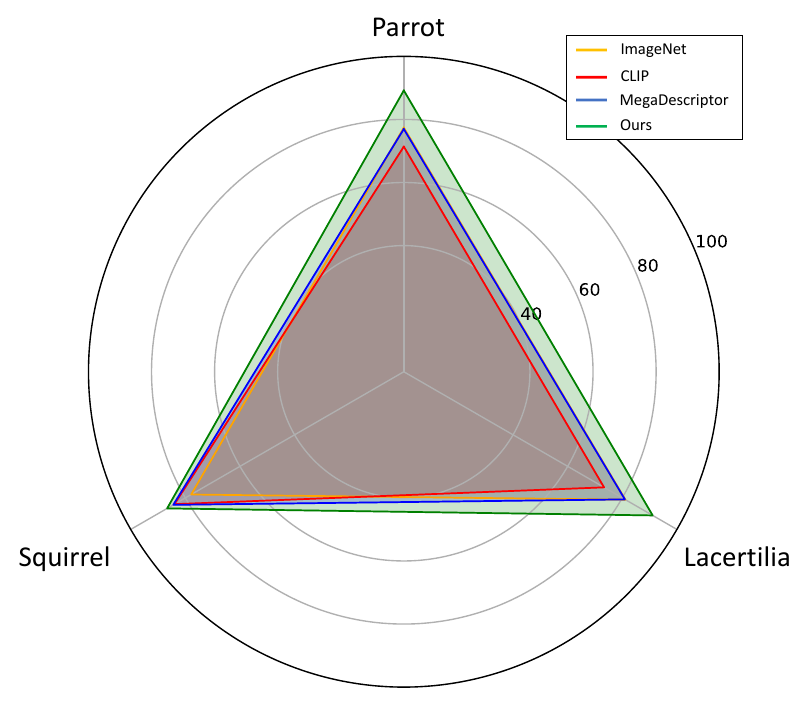}
            \caption{\textbf{Generality to unseen animal families.} Trained on our \datasetname achieves the best performance even for unseen animal families.}
            \label{fig:generality}
        \end{minipage}
    \end{tabular}
\end{figure}

\noindent\textbf{Fine-grained verification.}
In real world scenarios, animal verification is sometimes conducted within a specific breed rather than different breeds.
To meet this demand, we introduce fine-grained verification where positive and negative pairs are chosen only from the same breeds.
We refer to the top-10 dog breeds of individuals in our datasets: Chihuahua, Dachshund, French bulldog, Golden retriever, Miniature dachshund, Pomeranian, Shiba inu, Shih tzu, Toy poodle, and Yorkshire terrier.
We construct the fine-grained test sets and evaluate the models on each breed.
The results are visualized in Fig.~\ref{fig:fine_dog}.
It can be seen that our model achieves a good performance on all the breeds (98.30\% on average). 
The results achieved for the baseline models are lower in comparison (84.99\%, 80.78\%, and 86.25\% for ImageNet, CLIP, and MegaDescriptor, respectively). 
Moreover, we find that the results achieved with MegaDescriptor show significantly more degradation than the cross-breed verification result in Table~\ref{tb:verification} from 93.75\% to 86.25\%.
while our model maintains the results at a high-level (from 99.01\% to 98.39\%).
This result indicates that our dataset overcomes the limitation of the previous dog dataset~\cite{SMALST} that suffers from the poor performance for specific breed verification owing to its insufficient number of individuals (192 individuals).

\noindent\textbf{Generality to Unseen Families.}
Lastly, we evaluate the verification generality to unseen animal families. 
To achieve this, we additionally collect 100 identities each of Parrot, Lacertilia, and Squirrel, which are not contained in either our dataset or WildLifeDataset~\cite{WildlifeDatasets}. 
We compare our model that is jointly trained on our dataset across families with the baseline models in Fig.~\ref{fig:generality}.
Our model outperforms the baselines on all the species. 
For Parrot, particularly, our model achieves an AUC of 88.99\%, revealing its superiority over MegaDescriptor with a large margin. 
This result supports the generality and variety of our \datasetnameshort dataset.

\section{Conclusion}
We introduced the \datasetname, a comprehensive animal identification dataset encompassing 13 families, featuring 257,484 unique individuals across 319 breed categories. 
We have collected images and their information and conducted automated and manual filtering to ensure the quality of the large-scale dataset.
This dataset also includes detailed annotations of sex, breed, colors, and patterns to facilitate more investigation and application in a real-world scenario.
We establish two main benchmarks: 1) re-identification of seen individuals and 2) verification of unseen individuals. 

Our experiments show the generality of the models trained on our dataset to verification on other datasets or unseen animal families. 
We also found that there is still room for improvement in the integrated identification of multiple animal families. 
To promote further research, we will make this dataset, experiment code, and models available to the research communities. 
This dataset will enable computer vision researchers to tackle animal face identification across a wide range of breeds and push forward progress on animal face identification tasks.

\clearpage  

\section*{Acknowledgements}
We thank the Great Ape Information Network
(\url{http://www.shigen.nig.ac.jp/gain/}) for providing Chimpanzee face images. 
We also thank Dr. Takuma Yagi from AIST for comments on paper proofreading. 
Finally, we thank Prof. Toshihiko Yamasaki from the University of Tokyo for
providing computation resources.

%
%
\bibliographystyle{splncs04}
\bibliography{main}

\begin{thebibliography}{10}
\providecommand{\url}[1]{\texttt{#1}}
\providecommand{\urlprefix}{URL }
\providecommand{\doi}[1]{https://doi.org/#1}

\bibitem{centerloss_repo}
{Center Loss Pytorch}. \url{https://github.com/KaiyangZhou/pytorch-center-loss.git}

\bibitem{insightface_repo}
{InsightFace}. \url{https://github.com/deepinsight/insightface}

\bibitem{openclip}
Openclip. \url{https://doi.org/10.5281/zenodo.5143773}

\bibitem{torchvision2016}
Torchvision: Pytorch's computer vision library. \url{https://github.com/pytorch/vision}

\bibitem{tripletloss_repo}
{Triplet Loss Pytorch}. \url{https://www.kaggle.com/code/hirotaka0122/triplet-loss-with-pytorch}

\bibitem{BelugaID2022}
Beluga id 2022 (2022), \url{https://lila.science/datasets/beluga-id-2022/}

\bibitem{HyenaID2022}
Hyena id 2022 botswana predator conservation trust (2022). panthera pardus csv custom export. (2022), \url{https://lila.science/datasets/hyena-id-2022/}

\bibitem{LeopardID2022}
Leopard id 2022 botswana predator conservation trust (2022). panthera pardus csv custom export. (2022), \url{https://lila.science/datasets/leopard-id-2022/}

\bibitem{zindi}
Turtle recall: Conservation challenge (2022), \url{https://zindi.africa/competitions/turtle-recall-conservation-challenge}

\bibitem{seaturtle2022}
Adam, L., \v{C}erm\'ak, V., Papafitsoros, K., Picek, L.: Seaturtleid2022: A long-span dataset for reliable sea turtle re-identification. In: IEEE/CVF Winter Conference on Applications of Computer Vision (WACV). pp. 7146--7156 (2024)

\bibitem{OpenCows2020}
Andrew, W., Gao, J., Mullan, S., Campbell, N., Dowsey, A.W., Burghardt, T.: Visual identification of individual holstein-friesian cattle via deep metric learning. In: Computers and Electronics in Agriculture. vol.~185, p. 106133 (2021)

\bibitem{AerialCattle2017}
Andrew, W., Greatwood, C., Burghardt, T.: Visual localisation and individual identification of holstein friesian cattle via deep learning. In: IEEE/CVF International Conference on Computer Vision (ICCV) Workshops. pp. 2850--2859 (2017)

\bibitem{FriesianCattle2017}
Andrew, W., Greatwood, C., Burghardt, T.: Visual localisation and individual identification of holstein friesian cattle via deep learning. In: IEEE/CVF International Conference on Computer Vision (ICCV) Workshops. pp. 2850--2859 (2017)

\bibitem{FriesianCattle2015}
Andrew, W., Hannuna, S., Campbell, N., Burghardt, T.: Automatic individual holstein friesian cattle identification via selective local coat pattern matching in rgb-d imagery. In: International Conference on Image Processing (ICIP). pp. 484--488 (2016)

\bibitem{gorilla_zoo}
Brookes, O., Burghardt, T.: A dataset and application for facial recognition of individual gorillas in zoo environments. In: International Conference on Pattern Recognition(ICPR) (2020)

\bibitem{HappyWhale}
Cheeseman, T., Southerland, K., Park, J., Olio, M., Flynn, K., Calambokidis, J., Jones, L., Garrigue, C., Frisch-Jordán, A., Howard, A., Reade, W., Neilson, J., Gabriele, C., Clapham, P.: Advanced image recognition: a fully automated, high-accuracy photo-identification matching system for humpback whales. vol.~102, p. 915–929 (2021)

\bibitem{noaa-right-whale-recognition}
Christin B.~Khan, Shashank, W.K.: Right whale recognition (2015), \url{https://kaggle.com/competitions/noaa-right-whale-recognition}

\bibitem{Dahlborn2013ReportOT}
Dahlborn, K., Bugnon, P., Nevalainen, T., Raspa, M., Verbost, P.M., Spangenberg, E.M.F.: Report of the federation of european laboratory animal science associations working group on animal identification. vol.~47, pp. 2--11 (2013)

\bibitem{imagenet}
Deng, J., Dong, W., Socher, R., Li, L.J., Li, K., Fei-Fei, L.: Imagenet: A large-scale hierarchical image database. In: IEEE/CVF Conference on Computer Vision and Pattern Recognition (CVPR). pp. 248--255 (2009)

\bibitem{arcface}
Deng, J., Guo, J., Xue, N., Zafeiriou, S.: Arcface: Additive angular margin loss for deep face recognition. In: Proceedings of the IEEE/CVF Conference on Computer Vision and Pattern Recognition (CVPR). pp. 4690--4699 (2019)

\bibitem{LionData}
Dlamini, N., Zyl, T.L.v.: Automated identification of individuals in wildlife population using siamese neural networks. In: 2020 7th International Conference on Soft Computing \& Machine Intelligence (ISCMI). pp. 224--228 (2020)

\bibitem{vit}
Dosovitskiy, A., Beyer, L., Kolesnikov, A., Weissenborn, D., Zhai, X., Unterthiner, T., Dehghani, M., Minderer, M., Heigold, G., Gelly, S., Uszkoreit, J., Houlsby, N.: An image is worth 16x16 words: Transformers for image recognition at scale. The International Conference on Learning Representations (ICLR)  (2021)

\bibitem{BirdIndividualID}
Ferreira, A.C., Silva, L.R., Renna, F., Brandl, H.B., Renoult, J., Farine, D.R., Covas, R., Doutrelant, C.: {Deep learning-based methods for individual recognition in small birds}. In: {Methods in Ecology and Evolution}. vol.~11, pp. 1072--1085 (2020)

\bibitem{ctaiczoo}
Freytag, A., Rodner, E., Simon, M., Loos, A., K{\"u}hl, H.S., Denzler, J.: Chimpanzee faces in the wild: Log-euclidean cnns for predicting identities and attributes of primates. In: Rosenhahn, B., Andres, B. (eds.) Pattern Recognition. pp. 51--63. Springer International Publishing, Cham (2016)

\bibitem{Cows2021}
Gao, J., Burghardt, T., Andrew, W., Dowsey, A.W., Campbell, N.W.: Towards self-supervision for video identification of individual holstein-friesian cattle: The cows2021 dataset. arXiv preprint arXiv:2105.01938  (2021)

\bibitem{ms1m}
Guo, Y., Zhang, L., Hu, Y., He, X., Gao, J.: Ms-celeb-1m: A dataset and benchmark for large-scale face recognition. In: Leibe, B., Matas, J., Sebe, N., Welling, M. (eds.) European Conference on Computer Vision (ECCV). pp. 87--102. Springer International Publishing, Cham (2016)

\bibitem{Zebrafish}
Haurum, J.B., Karpova, A., Pedersen, M., Bengtson, S.H., Moeslund, T.B.: Re-identification of zebrafish using metric learning. In: IEEE Winter Conference on Applications of Computer Vision (WACV) Workshops. pp. 1--11 (2020)

\bibitem{resnet}
He, K., Zhang, X., Ren, S., Sun, J.: Deep residual learning for image recognition. In: IEEE/CVF Conference on Computer Vision and Pattern Recognition (CVPR). pp. 770--778 (2016)

\bibitem{MPDD}
He, Z., Qian, J., Yan, D., Wang, C., Xin, Y.: Animal re-identification algorithm for posture diversity. In: ICASSP 2023 - 2023 IEEE International Conference on Acoustics, Speech and Signal Processing (ICASSP). pp.~1--5 (2023)

\bibitem{biometric_sheep}
Hitelman, A., Edan, Y., Godo, A., Berenstein, R., Lepar, J., Halachmi, I.: Biometric identification of sheep via a machine-vision system. vol.~194, p. 106713 (2022)

\bibitem{tripletnet}
Hoffer, E., Ailon, N.: Deep metric learning using triplet network. In: arXiv (2018)

\bibitem{WhaleSharkID}
Holmberg, J., Norman, B., Arzoumanian, Z.: Estimating population size, structure, and residency time for whale sharks rhincodon typus through collaborative photo-identification. vol.~7, pp. 39 -- 53. Inter-Research Science Center (2009)

\bibitem{HumpbackWhaleID}
Howard, A., inversion, Southerland, K., Cheeseman, T.: Humpback whale identification challenge (2018), \url{https://kaggle.com/competitions/whale-categorization-playground}

\bibitem{lfw}
Huang, G.B., Ramesh, M., Berg, T., Learned-Miller, E.: Labeled faces in the wild: A database for studying face recognition in unconstrained environments. No. 07-49 (2007)

\bibitem{horse}
Jarraya, I., Ouarda, W., Alimi, A.M.: A preliminary investigation on horses recognition using facial texture features. In: IEEE International Conference on Systems, Man, and Cybernetics. pp. 2803--2808 (2015)

\bibitem{fishnet}
Khan, F.F., Li, X., Temple, A.J., Elhoseiny, M.: Fishnet: A large-scale dataset and benchmark for fish recognition, detection, and functional trait prediction. In: Proceedings of the IEEE/CVF International Conference on Computer Vision (ICCV). pp. 20496--20506 (October 2023)

\bibitem{animalweb}
Khan, M.H., McDonagh, J., Khan, S., Shahabuddin, M., Arora, A., Khan, F.S., Shao, L., Tzimiropoulos, G.: Animalweb: A large-scale hierarchical dataset of annotated animal faces. In: IEEE/CVF Conference on Computer Vision and Pattern Recognition (CVPR). pp. 6937--6946 (2020)

\bibitem{adaface}
Kim, M., Jain, A.K., Liu, X.: Adaface: Quality adaptive margin for face recognition. In: IEEE/CVF Conference on Computer Vision and Pattern Recognition (CVPR). pp. 18729--18738 (2022)

\bibitem{Cloudtag}
Krishnan, S., S., S.: Cloud iot based novel livestock monitoring and identification system using uid. In: Sensor Review. vol.~38, pp. 21--33 (2017)

\bibitem{anyface}
Kuzdeuov, A., Koishigarina, D., Varol, H.A.: Anyface: A data-centric approach for input-agnostic face detection. In: IEEE International Conference on Big Data and Smart Computing (BigComp). pp. 211--218 (2023)

\bibitem{StripeSpotter}
Lahiri, M., Tantipathananandh, C., Warungu, R., Rubenstein, D.I., Berger-Wolf, T.Y.: Biometric animal databases from field photographs: identification of individual zebra in the wild. In: Proceedings of the ACM International Conference on Multimedia Retrieval. Association for Computing Machinery (2011)

\bibitem{LESLIE201086}
Leslie, E., Hernández-Jover, M., Newman, R., Holyoake, P.: Assessment of acute pain experienced by piglets from ear tagging, ear notching and intraperitoneal injectable transponders. In: Applied Animal Behaviour Science. vol.~127, pp. 86--95 (2010)

\bibitem{ATRW}
Li, S., Li, J., Tang, H., Qian, R., Lin, W.: Atrw: A benchmark for amur tiger re-identification in the wild. In: Proceedings of the 28th ACM International Conference on Multimedia. p. 2590–2598. MM '20, Association for Computing Machinery, New York, NY, USA (2020)

\bibitem{sheep_transformer}
Li, X., Xiang, Y., Li, S.: Combining convolutional and vision transformer structures for sheep face recognition. In: Computers and Electronics in Agriculture. vol.~205, p. 107651 (2023)

\bibitem{sphereface}
Liu, W., Wen, Y., Yu, Z., Li, M., Raj, B., Song, L.: Sphereface: Deep hypersphere embedding for face recognition. In: IEEE/CVF Conference on Computer Vision and Pattern Recognition (CVPR). pp. 6738--6746 (2017)

\bibitem{swin}
Liu, Z., Lin, Y., Cao, Y., Hu, H., Wei, Y., Zhang, Z., Lin, S., Guo, B.: Swin transformer: Hierarchical vision transformer using shifted windows. In: IEEE/CVF International Conference on Computer Vision (ICCV). pp. 9992--10002 (2021)

\bibitem{Loos2013chimp}
Loos, A., Ernst, A.: An automated chimpanzee identification system using face detection and recognition. In: EURASIP Journal on Image and Video Processing. vol.~2013 (2013)

\bibitem{adaptive_pig}
Marsot, M., Mei, J., Shan, X., Ye, L., Feng, P., Yan, X., Li, C., Zhao, Y.: An adaptive pig face recognition approach using convolutional neural networks. In: Computers and Electronics in Agriculture. vol.~173, p. 105386 (2020)

\bibitem{catflw}
Martvel, G., Shimshoni, I., Zamansky, A.: Automated detection of cat facial landmarks. In: International Journal of Computer Vision (IJCV). pp. 1--16 (2023)

\bibitem{mouse_welfare}
Mazlan, N.H., Lopez~Salesansky, N., Burn, C., Wells, D.: Mouse identification methods and potential welfare issues: A survey of current practice in the uk. In: Animal Technology and Welfare. vol.~13, pp. 1--10 (2014)

\bibitem{Giraffes}
Miele, V., Dussert, G., Spataro, B., Chamaill{\'e}-Jammes, S., Allain{\'e}, D., Bonenfant, C.: {Revisiting animal photo-identification using deep metric learning and network analysis}. In: {Methods in Ecology and Evolution}. vol.~12, pp. 863--873. {Wiley} (2021)

\bibitem{flicker_dog}
Moreira, T., Perez, M., Werneck, R., Valle, E.: Where is my puppy? retrieving lost dogs by facial features. In: Multimedia Tools and Applications. vol.~76, p. 15325–15340 (2017)

\bibitem{agedb}
Moschoglou, S., Papaioannou, A., Sagonas, C., Deng, J., Kotsia, I., Zafeiriou, S.: Agedb: The first manually collected, in-the-wild age database. In: IEEE/CVF Conference on Computer Vision and Pattern Recognition (CVPR) Workshops. pp. 1997--2005 (2017)

\bibitem{DogFaceNet}
Mougeot, G., Li, D., Jia, S.: A deep learning approach for dog face verification and recognition. In: Nayak, A.C., Sharma, A. (eds.) PRICAI 2019: Trends in Artificial Intelligence. pp. 418--430. Springer International Publishing, Cham (2019)

\bibitem{Drosophila}
Murali, N., Taylor, G., Levine, J.: Can drosophila melanogaster tell who’s who? In: PLOS ONE. vol.~13, p. e0205043 (2018)

\bibitem{SealID}
Nepovinnykh, E., Eerola, T., Biard, V., Mutka, P., Niemi, M., Kunnasranta, M., Kälviäinen, H.: Sealid: Saimaa ringed seal re-identification dataset. Sensors  \textbf{22}(19) (2022)

\bibitem{seaturtleid}
Papafitsoros, K., Adam, L., Čermák, V., Picek, L.: Seaturtleid: A novel long-span dataset highlighting the importance of timestamps in wildlife re-identification. In: arXiv (2023)

\bibitem{GiraffeZebraID}
Parham, J., Crall, J., Stewart, C., Berger-Wolf, T., Rubenstein, D.: Animal population censusing at scale with citizen science and photographic identification. In: SS-17-01. pp. 37--44. AAAI Spring Symposium (2017)

\bibitem{clip}
Radford, A., Kim, J.W., Hallacy, C., Ramesh, A., Goh, G., Agarwal, S., Sastry, G., Askell, A., Mishkin, P., Clark, J., Krueger, G., Sutskever, I.: Learning transferable visual models from natural language supervision. In: arXiv (2021)

\bibitem{ROBERTS200618}
Roberts, C.: Radio frequency identification (rfid). In: Computers \& Security. vol.~25, pp. 18--26 (2006)

\bibitem{welfare_tatto}
Roughan, J., Sevenoaks, T.: Welfare and scientific considerations of tattooing and ear-tagging for mouse identification. In: Journal of the American Association for Laboratory Animal Science: JAALAS. pp. 142--153 (2018)

\bibitem{frontal}
Sengupta, S., Chen, J.C., Castillo, C., Patel, V.M., Chellappa, R., Jacobs, D.W.: Frontal to profile face verification in the wild. In: IEEE Winter Conference on Applications of Computer Vision (WACV). pp.~1--9 (2016)

\bibitem{animalface_eccv20}
Shi, X., Yang, C., Xia, X., Chai, X.: Deep cross-species feature learning for animal face recognition via residual interspecies equivariant network. In: European Conference on Computer Vision (ECCV). p. 667–682 (2020)

\bibitem{shinoda2024openanimaltracks}
Shinoda, R., Shiohara, K.: Openanimaltracks: A dataset for animal track recognition. arXiv preprint arXiv:2406.09647  (2024)

\bibitem{ndd20}
Trotter, C., Atkinson, G., Sharpe, M., Richardson, K., McGough, A.S., Wright, N., Burville, B., Berggren, P.: Ndd20: A large-scale few-shot dolphin dataset for coarse and fine-grained categorisation. In: arXiv (2020)

\bibitem{WildlifeDatasets}
\v{C}erm\'ak, V., Picek, L., Adam, L., Papafitsoros, K.: Wildlifedatasets: An open-source toolkit for animal re-identification. In: IEEE/CVF Winter Conference on Applications of Computer Vision (WACV). pp. 5953--5963 (2024)

\bibitem{SeaStarReID2023}
Wahltinez, O., Wahltinez, S.: An open‐source general purpose machine learning framework for individual animal re‐identification using few‐shot learning. In: Methods in Ecology and Evolution. vol.~15, pp. 373--387 (2024)

\bibitem{cosface}
Wang, H., Wang, Y., Zhou, Z., Ji, X., Gong, D., Zhou, J., Li, Z., Liu, W.: Cosface: Large margin cosine loss for deep face recognition. In: IEEE/CVF Conference on Computer Vision and Pattern Recognition (CVPR). pp. 5265--5274 (2018)

\bibitem{IPanda50}
Wang, L., Ding, R., Zhai, Y., Zhang, Q., Tang, W., Zheng, N., Hua, G.: Giant panda identification. IEEE Transactions on Image Processing  \textbf{30},  2837--2849 (2021)

\bibitem{center}
Wen, Y., Zhang, K., Li, Z., Qiao, Y.: A discriminative feature learning approach for deep face recognition. In: Leibe, B., Matas, J., Sebe, N., Welling, M. (eds.) European Conference on Computer Vision (ECCV). pp. 499--515 (2016)

\bibitem{macaques}
Witham, C.L.: Automated face recognition of rhesus macaques. In: Journal of Neuroscience Methods. vol.~300, pp. 157--165 (2018)

\bibitem{webface}
Zhu, Z., Huang, G., Deng, J., Ye, Y., Huang, J., Chen, X., Zhu, J., Yang, T., Lu, J., Du, D., Zhou, J.: Webface260m: A benchmark unveiling the power of million-scale deep face recognition. In: IEEE/CVF Conference on Computer Vision and Pattern Recognition (CVPR). pp. 10492--10502 (2021)

\bibitem{PolarBearVidID}
Zuerl, M., Dirauf, R., Koeferl, F., Steinlein, N., Sueskind, J., Zanca, D., Brehm, I., Fersen, L.v., Eskofier, B.: Polarbearvidid: A video-based re-identification benchmark dataset for polar bears. In: Animals. vol.~13 (2023)

\bibitem{SMALST}
Zuffi, S., Kanazawa, A., Berger-Wolf, T., Black, M.J.: Three-d safari: Learning to estimate zebra pose, shape, and texture from images “in the wild”. In: IEEE/CVF International Conference on Computer Vision (ICCV). pp. 5358--5367 (2019)

\end{thebibliography}

\clearpage
\appendix

\section{Fine-grained Annotations}
\begin{figure}[tb]
  \centering
  \includegraphics[height=12cm]{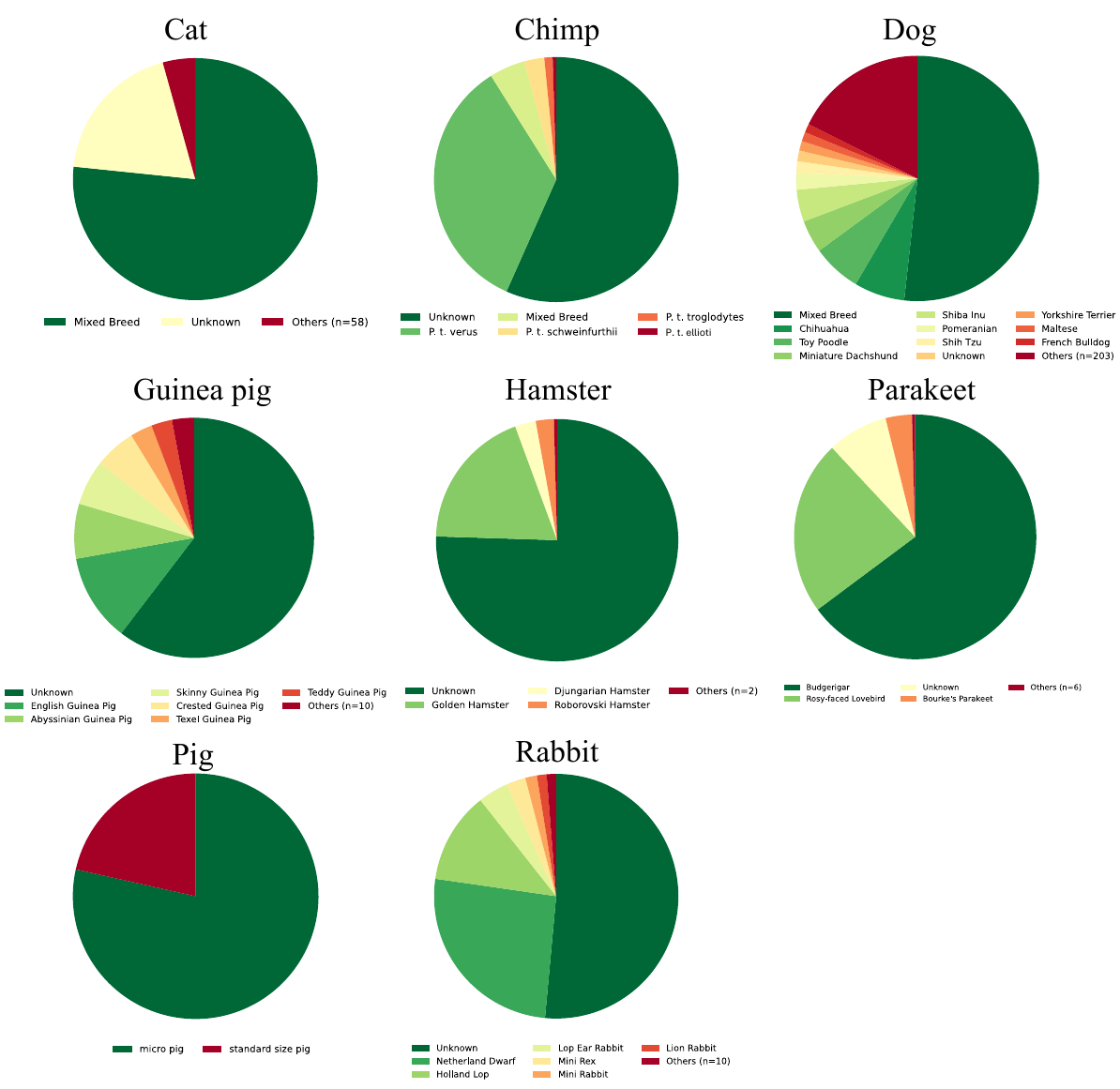}
  \caption{\textbf{Breed annotations.} }
  \label{fig:example_breed}
\end{figure}

\begin{figure}[tb]
  \centering
  \includegraphics[height=16cm]{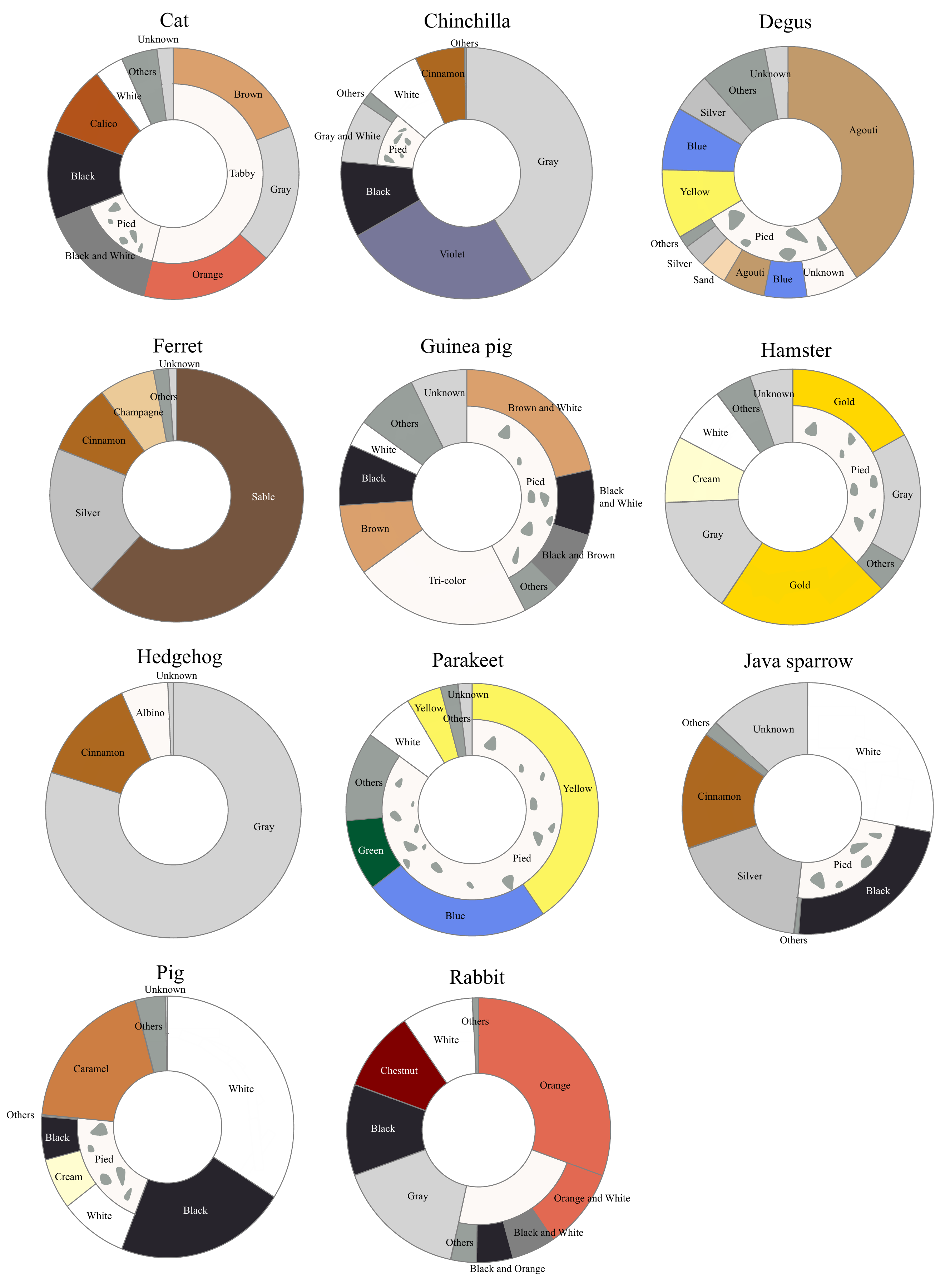}
  \caption{\textbf{Color and Pattern annotations.} }
  \label{fig:example_colors}
\end{figure}

We list breed annotations statistics for eight animal classes (Cat, Chimp, Dog, Guinea pig, Hamster, Parakeet, Pig, and Rabbit) in Fig~\ref{fig:example_breed}. 
Similarly, color and pattern annotations statistics for eleven animal classes (Cat, Chinchilla, Degus, Ferret, Guinea pig, Hamster, Hedgehog, Parakeet, Java sparrow, Pig, and Rabbit) are presented in Fig~\ref{fig:example_colors}. 
Our dataset provides color and pattern information through two-tier hierarchical annotations, especially for animal families whose colors commonly distinguish individuals.

\section{Implementation Details}

\noindent\textbf{Pre-processing.}
We input the images with 224$\times$224 pixels for our models.
We use horizontal flip to augment the training images.
For the other models, we strictly follow the defined pre-processings.

\noindent\textbf{Triplet loss.}
We use a 3rd-party implementation~\cite{tripletloss_repo}.
We set the margin parameter to 0.5.

\noindent\textbf{Center loss.}
We use a 3rd-party implementation~\cite{centerloss_repo}.

\noindent\textbf{ArcFace loss.}
We use a 3rd-party implementation~\cite{insightface_repo}.
We set the margin and scale parameters to 0.5 and $\sqrt{2}\log(C-1)$ where $C$ denotes the number of individuals, respectively.

\noindent\textbf{ImageNet.}
We use the torchvision version~\cite{torchvision2016}.

\noindent\textbf{CLIP.}
We use the open clip version~\cite{openclip}.

\noindent\textbf{MegaDescriptor.}
We use the official implementation~\cite{WildlifeDatasets}.

\end{document}